# Endless Loops: Detecting and Animating Periodic Patterns in Still Images

TAVI HALPERIN, Lightricks, Israel
HANIT HAKIM, Lightricks, Israel
ORESTIS VANTZOS, Lightricks, Greece
GERSHON HOCHMAN, Lightricks, Israel
NETAI BENAIM, Lightricks, Israel
LIOR SASSY, Lightricks, Israel
MICHAEL KUPCHIK, Lightricks, Israel
OFIR BIBI, Lightricks, Israel
OHAD FRIED, Interdisciplinary Center, Herzliya, Israel

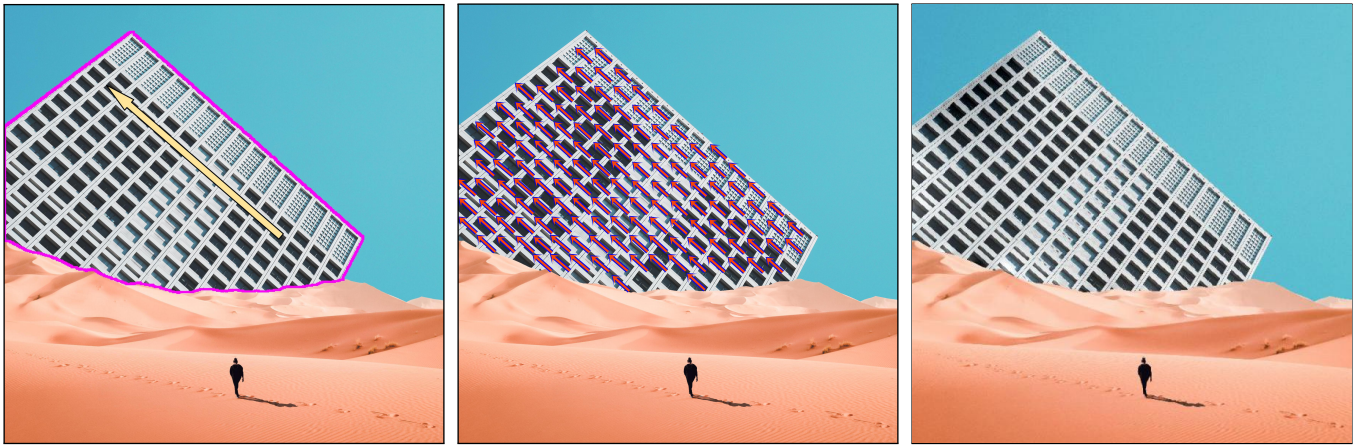

Fig. 1. *Creating a cinemagraph from a still image.* We look for self-similarities in the user-selected area of interest, solve for a displacement field, and use it to warp the image into a seamlessly looping animation; in this case each window smoothly sliding into the next one along the face of the building, a surreal yet aesthetically pleasing visual effect. **Please open in Acrobat Reader to see embedded videos.** Image credit: © Gal Nachmana

We present an algorithm for producing a seamless animated loop from a single image. The algorithm detects periodic structures, such as the windows of a building or the steps of a staircase, and generates a non-trivial displacement vector field that maps each segment of the structure onto a neighboring segment along a user- or auto-selected main direction of motion. This displacement field is used, together with suitable temporal and spatial smoothing, to warp the image and produce the frames of a continuous animation loop. Our cinemagraphs are created in under a second on a mobile device. Over 140,000 users downloaded our app and exported over 350,000 cinemagraphs. Moreover, we conducted two user studies that show that users prefer our method for creating surreal and structured cinemagraphs compared to more manual approaches and compared to previous methods.

Authors' addresses: Tavi Halperin, Lightricks, Israel, tavi@lightricks.com; Hanit Hakim, Lightricks, Israel, hanit@lightricks.com; Orestis Vantzos, Lightricks, Greece, orestis@lightricks.com; Gershon Hochman, Lightricks, Israel, gershon@lightricks; Netai Benaim, Lightricks, Israel, netai@lightricks.com; Lior Sassy, Lightricks, Israel, lsassy@lightricks.com; Michael Kupchik, Lightricks, Israel, michael@lightricks.com; Ofir Bibi, Lightricks, Israel, ofir@lightricks.com; Ohad Fried, Interdisciplinary Center, Herzliya, Israel, ofried@idc.ac.il.

## 1 INTRODUCTION

Ever since the dawn of photography people have been fascinated by photos and videos that appear realistic, yet contain a fantastical element. Early examples include *Lincoln's Ghost* [Mumler 1872], *Dirigible Docked on Empire State Building* [Unknown 1930a], and *Man on Rooftop with Eleven Men in Formation on His Shoulders* [Unknown 1930b]. These examples capture our imagination, as they combine the mundane and familiar with unexpected juxtaposition. While a dirigible airship and a building are perfectly ordinary, a photo-montage of the airship docked to the building is surprising and interesting. With the advent of image-centric social networks like *Instagram*, manipulated images and animations of this type are becoming quite popular and are widely shared.

One specific type of surreal manipulation is called a *cinemagraph* — a video in which most areas are static, and some areas contain seamless repeating motion. Cinemagraphs exist on the boundary between photos and videos, and have been used both by amateurs, and in professional news, advertisements, and fashion photography[1]. Also referred to as video textures or live photos, various techniques have been proposed for the creation of cinemagraphs and their variations [Endo et al. 2019; Holynski et al. 2020; Joshi et al. 2012; Liao et al. 2015, 2013; Schödl et al. 2000; Tompkin et al. 2011; Yeh and Li 2012]. In this work we tackle the problem of cinemagraph creation, with two distinct conditions:

(1) The input is a single image rather than a video.
(2) The desired output is a surreal visual such as a moving building.

While the first condition has been tackled in the past [Endo et al. 2019; Holynski et al. 2020], the combination of the two poses unique challenges, such as the inability to trivially use video datasets for training (since, for instance, buildings do not usually move in the real world).

Converting a single image into a video is an active area of research. Some techniques focus on natural physical phenomena such as waterfalls or smoke [Endo et al. 2019; Holynski et al. 2020; Okabe et al. 2009, 2018], or modeling the effect of wind on natural scenery [Chuang et al. 2005]. Others render a video from a virtual camera panning around the scene [Kopf et al. 2020; Niklaus et al. 2019; Shih et al. 2020], transfer motion from a video to a photo [Hornung et al. 2007; Tesfaldet et al. 2018; Weng et al. 2019], generate looping animal motion from an image [Xu et al. 2008], or generate sequences of perturbations of a single image [Rott Shaham et al. 2019]. The common denominator of these is that they reproduce (or re-imagine) a *realistic* video, and therefore have the privilege of harnessing datasets of real images and videos by extracting optical flow or to train a data driven approach. As mentioned above, this work aims at rendering a more fantastical, often unrealistic motion in the scene. Thus, we cannot easily exploit a corpus of real footage to train a learning based algorithm. Instead, we turned to a hybrid approach, combining a *conditional random field* (CRF) formulation with *deep pixel descriptors*.

Our approach exploits self similarity of image patches to generate a motion field, which is later used to render the looping video. The definition of self similarity in images is usually task dependent, and we found existing algorithms either too restrictive in what is considered a repetition [Pritts et al. 2014], or too slow for our needs [Aiger et al. 2012; Lukáč et al. 2017]. In contrast, we want an algorithm that is able to find patterns when they exist, but is flexible, allowing the generation of motion even for less repetitive textures. We also want the algorithm to be efficient enough to run at interactive rates on a mobile phone.

An ideal system would excel for all types of input images, including natural scenes (e.g. waterfalls) and human-made objects (e.g. buildings). As we show in Section 5, our system is more domain specific, achieving state-of-the-art results for images with repeating patterns, and reasonable (but inferior to state-of-the-art) results on less structured images. Nevertheless, our users report satisfaction from all our results, and we leave the development of a unified framework which excels on all domains as future work.

Our main contributions are:

(1) A fast two stage algorithm for detecting repeating patterns suitable for animated motion. The first stage efficiently solves a reduced 1D matching problem, and the second stage is extending that result to 2D using a CRF formulation (Section 3.1.1).
(2) A system that generates looping videos from a single image, that can create surreal motion which cannot be observed in datasets of real video (Section 3.2).
(3) An interactive mobile app for cinemagraph creation, usable by novices and experts alike (Section 5.3).

Moreover, we provide quantitative evidence for the ease of use and aesthetically pleasing results of the system, in the form of an extensive user satisfaction evaluation, including over 140,000 users.

## 2 RELATED WORK

Our work is related to pattern recognition, motion estimation, and cinemagraph creation.

### 2.1 Cinemagraphs

Also called *video textures* [Agarwala et al. 2005; Schödl et al. 2000] or *cliplets* [Joshi et al. 2012], cinemagraphs aim to produce looping videos by manipulating an input video, and potentially freezing parts of the frame. Manually creating cinemagraphs is time-consuming even for professional artists. In the computational photography community several methods were introduced to automate this process [Liao et al. 2013; Tompkin et al. 2011; Yeh and Li 2012]. Usually these methods take as input a video (often recorded with a static camera), and re-sample the frames in time to produce a seamless loop. Some focus on specific scenarios such as objects which undergo large motion [Bai et al. 2012] or portrait videos [Bai et al. 2013]. The input video may contain undesired motion in some areas of the frame, and thus some methods also freeze the areas which are not part of the main object motion, which adds to the perceptual experience of cinemagraphs. In a recent approach Holynski et al. [2020] trained a neural network to generate looping videos from still images, generating impressive videos of flowing waterfalls and other types of natural motion. They achieved photo-realistic results by training on real videos, formulating the problem similarly to frame interpolation [Niklaus and Liu 2020]. In this work we are aiming for a similar end goal, but with one major difference: we want a method that does not require any video supervision, thus making it applicable to a broad range of scenes without the need for expensive dataset collection for each video category. Moreover, we aim for a method that can also create *surreal* videos which do not occur in the real world, and thus cannot be learned from a dataset of real videos. We opted therefore to formulate the problem as an image pattern recognition task instead.

---

[1]https://cinemagraphs.com

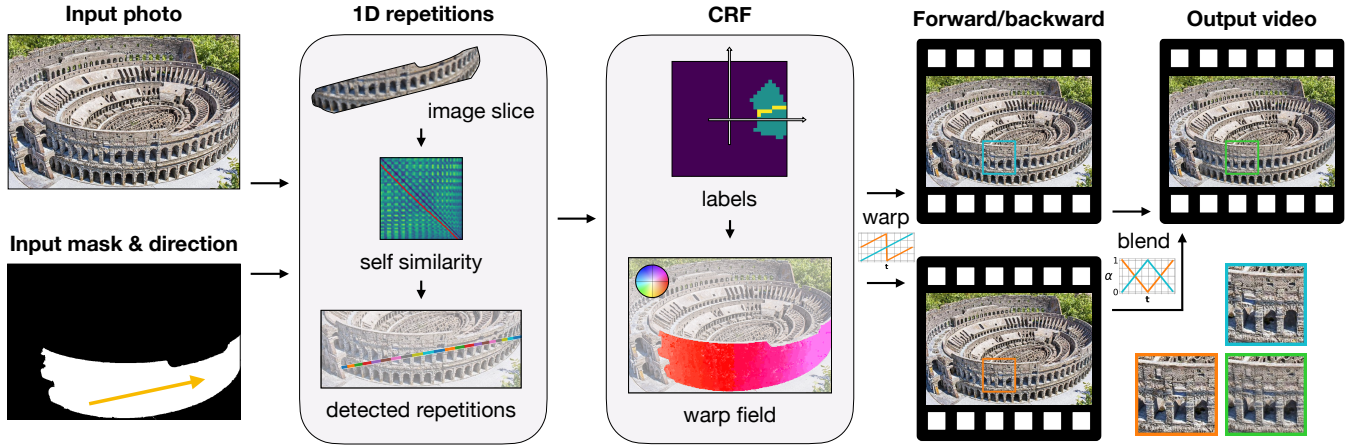

Fig. 2. *Cinemagraph animation pipeline.* Given the input image, mask and direction, we detect repetitions along a 1D curve **(1st stage)**, and use them to precompute a set of displacement vectors; we then solve a CRF where each pixel in the mask is assigned one of these vectors **(2nd stage)**. The frames of the seamlessly looping cinemagraph are produced by warping the image forwards and backwards and blending. Image credits: © canbedone

### 2.2 Pattern Recognition

Our method shares components with image completion methods, specifically those who use global optimization [Komodakis and Tziritas 2007; Pritch et al. 2009]. These use *Markov random fields* (MRF) to optimize offset values for existing pixels, in order to fill in the unknown pixels. Similarly, we use dense *conditional random fields* (CRF) to globally solve for pixel offsets. However, we overcome the main drawback of these types of methods, namely long running times, by restricting the labels (possible offsets) to a small set, carefully chosen in the first part of our two-step approach. The method of He and Sun [2012] computes patch similarity statistics in an image, and the offsets which appear the most are used to complete the unknown regions. We use a similar principle to automatically identify prominent directions of structure repetition, and suggest them to the user as candidate directions for motion generation (Section 4.4).

Unlike in the case of image completion, where pixels from distinct areas can be used to fill neighboring locations, we seek to evaluate a single smooth displacement field over the user-selected parts of the image that are to be animated. This property of our method shares similarities with global methods aimed at the discovery of repeating textures or symmetries in images [Aiger et al. 2012; Hays et al. 2006; Lee and Liu 2011; Liu et al. 2015; Lukáč et al. 2017; Park et al. 2009; Pritts et al. 2014; Zhang et al. 2012]. The common goal is to detect repeating textures (*texels*) together with the transformations that create them. Often, the underlying structure is a lattice of texels, but some methods also handle sparser repetitions and non lattice-based symmetries like reflection and glide-reflection. Our work is more connected to the former type of structure, as in our formulation the CRF step solves a problem related to the recovery of a near-regular pattern with translational symmetry [Liu et al. 2004], but we also indirectly allow for rotational symmetry via multiple directions (Section 4.2). By restricting ourselves to specific symmetries and by applying a two stage approach we perform the pattern recognition step fast enough for an interactive mobile app.

## 3 THE ALGORITHM

To render a seamlessly looping video, we need to first identify repeated structures in the given image, such as a staircase or a wall with tiles. Our algorithm takes as input

(a) an image to be animated,
(b) a binary mask indicating which pixels are allowed to move,
(c) a desired general direction of motion

and consists of the following steps (Figure 2):

(1) Detect image repetitions along a 1D curve, centered in the mask and aligned with the specified direction of motion (Section 3.1.1).
(2) Calculate a fixed set of displacement vectors from the offsets obtained in (1), to act as *labels* for a CRF. Solve the CRF to assign a suitable vector for each pixel inside the mask (Section 3.1.2).
(3) For each frame of the output video, perform two warps of the input image, one a suitable distance along the calculated displacement field and a second one along its inverse, and blend (Section 3.2).

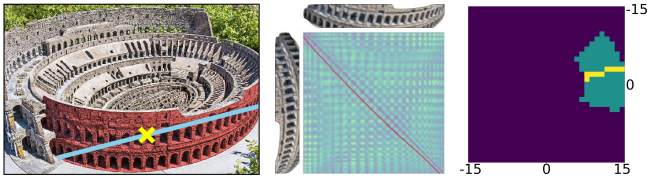

Fig. 3. **CRF labels. Left:** We extract an image slice around the center of mass of the given mask (yellow x) in the given direction (blue line). **Center:** We compute a self-similarity matrix from the slice of pixels. We find the shortest path in the matrix using dynamic programming (red line), and use its offsets from the main diagonal as a seed set of displacement vectors. **Right:** We rotate the seed set (yellow) to match the desired direction of motion, and dilate it radially to equip the CRF with a wider angular range of labels (turquoise). Image credits: © canbedone

### 3.1 Calculating the Displacement Field

The goal of the early stages of the pipeline is to calculate per-pixel 2D displacements between similar image regions, which can then be used to generate a plausible looking periodic motion. Our key observation is that the displacement fields that are suitable for cinemagraph creation can be usually constructed from a limited set of discrete displacement vectors, namely a main direction vector together with a small number of perturbations. Keeping this in mind, we split the displacement assignment problem into two sub-tasks. First, we solve an easier problem in 1D, identifying suitable displacement vectors along a single curve. Subsequently, these vectors are used to determine a small set of candidate 2D displacement vectors, that serve as the labels of a dense CRF problem, to assign one such displacement to each active pixel in the image. The limited number of labels is indeed key, as the running time of the dense CRF solver is quadratic in it.

*3.1.1 1D Repetitions.* Since the main direction of the motion is (loosely) constrained by the user, it makes sense to look first for image repetitions along that direction and within the region of the mask. Specifically, we focus on a line segment through the center of mass of the mask along the main direction, and sample a wide band of pixels around it (Figure 3, left).

The sample is treated as a sequence of column vectors $c_i$ of length equal to the width of the band, ordered from the beginning of the segment to its end. The objective is to match successive occurrences of the minimal unit of repetition (window, tile, stair, etc.) along the sample. We solve this using a dynamic programming formulation. We first build a self-similarity matrix $D_{ij} = \|c_i - c_j\|$ between the columns $c_i$ (Figure 3, center), and look for a path $(i_1, j_1), (i_2, j_2), \ldots$ spanning the main diagonal of the matrix that minimizes the total cost $\sum_k D_{i_k j_k}$ subject to the following constraints:

(a) *Continuity:* successive points on the path are 8-connected, $|i_k - i_{k+1}| \leq 1$ (likewise for $j$).
(b) *Monotonicity:* the indices of the points are non-decreasing, $i_k \leq i_{k+1}$ (likewise for $j$).
(c) *Bounded slope:* the path can not stay on the same matrix row or column for more than two entries, $i_{k-1} \neq i_{k+1}$ (likewise for $j$).

In typical dynamic programming fashion, we start from the top-left corner and calculate at each location of the matrix the cost of the cheapest path up to that location. Because of the continuity and monotonicity constraints, the calculation proceeds in a well-ordered manner from the top-left corner to the bottom and right edges of the matrix. With some extra book-keeping, storing at each location separately the cost of the cheapest path that arrives at the location from each of the three possible directions, we can restrict the calculation to only include paths that also satisfy the bounded slope condition.

The main technical challenge comes from the fact that the problem, as specified above, admits as a trivial solution the path $(1, 1)$, $(2, 2), \ldots$ that proceeds directly down the main diagonal with a total cost of 0. This solution is useless to us, since it corresponds to no motion whatsoever! We wish instead to find a non-trivial path of minimal cost, which we expect to correspond to a match between successive similar segments of the sample. To this end, we modify the self-similarity matrix so that nearby columns are artificially considered highly *dissimilar*:

$$D_{ij} := \begin{cases} \|c_i - c_j\|, & |i - j| > e \\ \infty, & |i - j| \leq e \end{cases}. \quad (1)$$

In our code we take $e = 8$.

As a further complication, we found that pinning the beginning and the end of the path at the top-left and bottom-right corners of the matrix respectively results in attenuated motion in those areas due to our dynamic programming constraints. We solve this by extending, via reflections, the self-similarity matrix, and using the extended matrix to calculate a continuous minimal path that spans the length of the sample twice. We take the middle part of the extended path, which is free from the strict initial/final conditions, and can be used to generate a more accurate motion.

*3.1.2 Displacement Assignment with CRF.* At this stage our goal is to determine a dense vector field, by assigning a displacement vector to each pixel inside the area to be animated. This assignment should be close to the 1D assignment, but also more flexible, while keeping relative smoothness. To this end, we formulate the problem as a global CRF optimization and encode our constraints as a suitable combination of unary and pair-wise potentials. We collect the various motion vectors we obtained from solving for 1D displacements, every $(i, j)$ pair in the optimal path corresponding to a displacement vector along the sampled curve, and use them as the set of labels for CRF. Since all displacement vectors are in the same direction, along the 1D line, we want to allow more angular freedom when assigning motion vectors to other pixels. We allow deviation from this constrained set of labels by introducing additional labels with angular deviation of up to 30 degrees from an existing label (Figure 3). We make further use of the 1D displacement vectors, by (nearest-neighbor) extrapolating them from the sampled line out to the rest of the image, assigning to each pixel $p_i$ an initial displacement vector $g_i$.

We use a dense CRF solver [Krähenbühl and Koltun 2013] to minimize an error function constructed from a unary and a pair-wise term. We use the notation introduced by Krähenbühl and Koltun. Every pixel $p_i$ is associated with a random variable $x_i$, taking values from the set of labels, and the objective is to minimize the following error function:

$$E(X) = \sum_i \psi_u(x_i) + \sum_{i,j} \psi_p(x_i, x_j) \quad (2)$$

The unary data term

$$\psi_u(x_i) := \|F_{p_i} - F_{p_i+x_i}\|^2 \left(1 + \lambda \left| \|x_i\| - \|g_i\| \right| \right) \quad (3)$$

is a perceptual distance between each pixel $p_i$ and its displacement. We use the first two convolutional layers (before downsampling) of VGG-16 [Simonyan and Zisserman 2014] as a *feature extractor*, concatenated to a 128-dimensional vector $F_{p_i}$ for every pixel. The perceptual distance is weighted with a factor that penalizes the displacement vector for straying too far from the initial guess $g_i$; because we do want to allow angular deviations from the main direction, we only take the difference of the norms. The constant $\lambda$ is set experimentally to 0.1.

The pair-wise potential

$$\psi_p(x_i, x_j) := \mu(x_i, x_j) \, exp\left(-\frac{\|p_i - p_j\|^2}{2\theta_\alpha^2} - \frac{\|m_i - m_j\|^2}{2\theta_\beta^2}\right) \quad (4)$$

promotes smoothness, by measuring the similarity between the labels of nearby pixels. The exponential bilateral weight vanishes quickly as the distance $\|p_i - p_j\|$ between the pixels increases beyond $\theta_\alpha = 10$; it also becomes essentially zero if the pixels have different mask values $m_i$ and $m_j$, as we set $\theta_\beta = 10^{-5} \ll 1$, essentially disconnecting the different components of the mask from each other. Solving the exponential bilateral weight formulation allows us to utilize the dense CRF solver [Krähenbühl and Koltun 2013]. Note that this effective weighting could be achieved instead by multiplying with the factor $[m_i = m_j]$, or manually excluding the pixels outside the mask from the sums in the error function.

The factor $\mu(x_i, x_j)$ is the *label compatibility function*. A common choice for this function for image segmentation tasks is the Potts model $\mu(x_i, x_j) = [x_i \neq x_j]$. However, in our case there are other more meaningful geometrical relations between the labels, which are displacement vectors. We use the *cosine similarity measure* raised to the 1/2 power

$$\mu(x_i, x_j) := \left(\frac{x_i \cdot x_j}{\|x_i\| \|x_j\|}\right)^{1/2}. \quad (5)$$

The exponent 1/2 produced less noisy labeling in our experiments. Note that the set of displacement vectors spans a limited range of angles by construction, hence the dot product in the definition is always positive. Given that the cosine similarity ignores the length of displacement vectors, we also tested using an additional term for length difference but did not see any meaningful improvement.

*3.1.3 Post-processing the Displacement Field.* Despite the presence of the smoothness-promoting pair-wise term in the error function, the raw label assignment produced by the CRF is still quite noisy. This is typical of the CRF method unless the number of labels is very small, and can not be rectified by running more iterations of the solver (Figure 4, bottom-left). Moreover, to generate motion from the labels we use backward warping according to the flow

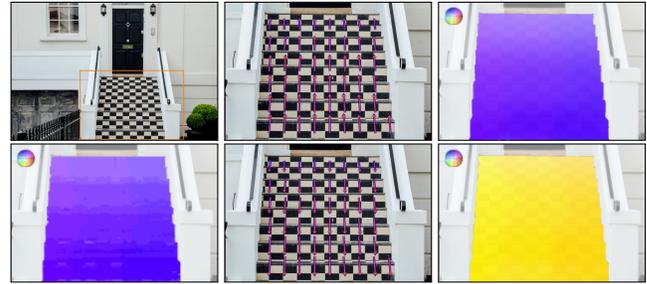

Fig. 4. *Reverse flow field.* **Left:** Image crop and flow from raw CRF output. **Center:** Sparsified and smoothed flow, trivial to invert. **Right:** Smooth dense forward and backward flow interpolated from the sparse set of flow vectors. We utilize these dense flow fields to warp the image to any timestamp in $[-1, 1]$. Image credit: © Evelyn Paris

(Section 3.2), so we want our displacement field to be invertible, which is unlikely for the field we get from the CRF.

To achieve a sufficiently smooth flow we apply a Gaussian kernel and subsample to get a sparse field; to densify it, we then interpolate the flow to all pixels using a *polyharmonic spline* with radial basis function $\phi(r) = r$. By inverting the sparse set of vectors and then densifying with the polyharmonic spline, we arrive at a smooth dense field that serves as a good approximation of the inverse flow, without the holes or collisions that we would get from directly inverting the displacement vectors of the raw CRF output (Figure 4).

### 3.2 Generation of Video Frames

Given the forward $F_1$ and inverted $F_2$ smooth displacement fields from the previous section, we wish to generate animation frames $I_t$ for a sequence of time stamps $t_0 = -1, \ldots, t_K = 1$. We require that for $t = -1$ and $t = 1$ the generated frame is exactly the input image $I$ so that the animation loops seamlessly. Let us denote the operation of warping an image $I$ by a displacement field $F$ as $warp(I; F)$. Using for stability the forward field for positive time stamps and the inverted one for negative, and setting $V_0 = I$, we consider the following animation $V_t$, $t \in [-1, 1]$:

$$V_t = \begin{cases} warp(I; t\, F_1), & t \geq 0 \\ warp(I; |t|\, F_2), & t < 0 \end{cases} \quad (6)$$

For a perfectly periodic input image $I$ and ideal displacement field $F_1$ and its inverse $F_2$, $V_{-1} = V_0 = V_1$ and this animation would loop seamlessly. In practice, $V_{-1}$ and $V_1$ are not exactly equal but rather similar to the input image, as the periodic patterns are shifted backwards/forwards by one unit of repetition. We take advantage of this similarity to generate the final seamless looping animation $I_t$, by alpha-blending the animation $V_t$ with a time-shifted version of itself:

$$I_t = \alpha(t) * V_t + (1 - \alpha(t)) * V_{\sigma(t)} \quad (7)$$

The piece-wise linear shift function $\sigma(t)$ can be seen plotted in orange in the *warp inset* of Figure 2, and likewise the alpha-blend factor $\alpha(t)$ is plotted in turquoise in the *blend inset* of Figure 2. They are tuned so that:

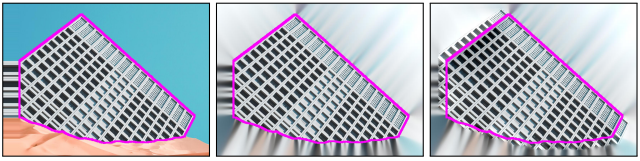

Fig. 5. *Extrapolating beyond the mask.* When displacement vectors point to pixels outside the mask, these pixels might end up blended into the moving part of the frame, creating artifacts (*color bleed*). To avoid this, we create an auxiliary texture that uses only the information inside the masked part of the image. **Left:** We pad the image to allow pulling out-of-bounds pixels. **Center:** We prepare the canvas by extrapolating to pixels outside the mask using an inpainting algorithm. **Right:** Once we compute the flow, we warp from inside the border of the mask outwards, and Poisson blend with the texture. Notice how the facade of the building extends outside the mask. Image credit: © Gal Nachmana

(a) The frames that are blended together from the two animations are always similar (corresponding approximately to a unit shift for any periodic image patterns).
(b) Each animation is multiplied with a zero blend factor exactly at the time of its discontinuity.

The first property minimizes any blending artifacts, and the second eliminates the jump discontinuities.

When warping, special care should be taken to handle pixels whose source or target is an unmasked pixel. In that case we don't want pixels from unrelated objects to flow into our masked region (might also happen when the unmasked pixel is the target of displacement for $t < 0$). We create a new texture to use in the warping instead of the original image, by extending the repeating pattern beyond the edges of the mask. The part of the new texture inside the mask, which is identical to the input image, is first extrapolated to the rest of the texture using inpainting [Telea 2004]. We then warp from inside the mask to the outside with both forward and backward flows and clone that into the inpainted texture with *Poisson blending* [Pérez et al. 2003] (see Figure 5). Note that we perform inpainting and Poisson blending once, and warp this processed image to generate all frames of the output video.

## 4 EXTENSIONS

Thus far we have considered motions restricted to one connected masked region, with one principle motion direction selected for the entire image. It would be useful, however, to be able to generate endless loops with multiple active regions (either connected or disconnected) that individually move in different directions. One other interesting case is when a video is given as input instead of a still image; the challenge then is to propagate the generated motion from the initial frame to the rest of the video without introducing artifacts. We also consider in this section the option of estimating the main direction of motion automatically, instead of receiving it as input from the user. Finally, we discuss letting the motion field vanish smoothly towards the edges of the mask, instead of a hard cut-off, which yields better results for certain natural scenes.

### 4.1 Multiple Disconnected Masks

Since the directional input is not necessarily associated with a spatial location, we might wish to apply a single input direction to multiple masks simultaneously. Running the entire algorithm on each mask independently usually improves results over computing a single set of CRF labels, as the pattern inside each mask might have a different orientation and/or periodicity. We show examples for these cases in the supplementary material.

### 4.2 Multiple Directions

Assigning individual directions of motion to multiple mutually-disconnected areas is equivalent to running multiple independent copies of the algorithm. On the other hand, moving different parts of a single connected mask in different directions (to animate billowing smoke for instance) is not as simple. In this case the user input takes the form of directions anchored to specific image locations. We split the mask into *Voronoi cells* centered at these locations and independently run the algorithm to obtain a CRF-generated displacement field on each one of these cells. We then follow the procedure of Section 3.1.3 to combine the individual fields into a single smooth dense displacement field, that can be used to animate the image.

In our interactive app, we allow the user to express their intended motion by freely drawing one or more strokes (curves) as guidelines. If a stroke is highly curved we break it internally into multiple smaller line segments, and proceed with running the method as described above.

### 4.3 Video as Input

Although our algorithm operates on a single image, we can apply the computed motion field on every frame of a video. To make the output consistent with the motion in the video we estimate a planar homography transformation of the masked pixels from the first frame of the input video to all other frames, and use these homographies to warp the moving object in future frames. To get highly accurate homographies we use the *restart tracking scheme* introduced by Halperin et al. [2019]. We show an example of introducing motion to video in the supplementary video. We tested a more complicated approach, which intuitively should perform better in the presence of exposure changes, namely warping the displacement fields to future frames and using those to generate the frames themselves. However, this approach resulted in flicker artifacts near the boundary of the mask, because the pixels that are pulled from inpainted texture outside the mask are not consistent throughout the video.

### 4.4 Automatic Estimation of the Main Direction

Without user input, the problem of identifying the main direction of motion is ill-defined, even when there is a natural direction of repetitions in the image. For example, for a given direction of motion the opposite direction is equally likely. We propose a simple technique to automatically suggest a motion direction. We observe that an optimal direction of motion is characterised by strong patch similarity along itself. Taking advantage of this, we extract a descriptor $F_p$ for every pixel $p$ (with the same features used to compute

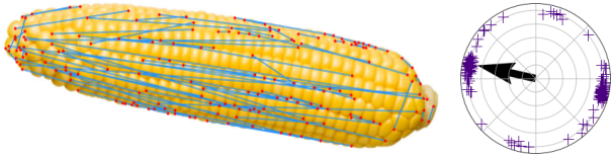

Fig. 6. *Automatic direction of motion.* **(Left)** We compute 'best buddy' pairs between selected points, using a local feature descriptor. **(Right)** We use these pairs to select a single global direction by vote. Note that a direction and its opposite always have the same number of votes and we choose one of them arbitrarily. Image credits: © Tracy Decourcy

the unary potentials of the CRF in Section 3.1.2) and use it to compute bidirectional nearest neighbours, a procedure also known as *best buddies similarity* [Dekel et al. 2015]. Each pair of best buddies votes in favour of the direction of their offset, and the direction with the most votes is chosen. This step is similar to the method of He and Sun [2012], where they find global image translations by local voting for offsets. Contrary to their method, which aims to find exact offsets, we are interested only in the direction of the offsets. Therefore, after choosing the offset with highest support, we eliminate all votes in the same (and opposite) directions, before voting for the next direction. We limit the number of suggested directions to three. In practice we do not use the descriptors of every single pixel, for efficiency and because including non distinctive pixels has negative effect on overall accuracy. Instead, we sample a fixed number of corner points [Shi et al. 1994]. The voting scheme is visualized in Figure 6.

### 4.5 Gradual Motion Attenuation Near the Mask Boundary

We observe in many images that *hard masking*, where the motion field drops sharply to zero at the boundary of the mask, generates a visible motion edge. This behaviour is actually desirable for man-made objects, where the boundary of the object is sharp and we don't want pixels outside of the object to move. However in other scenes, such as natural landscapes, it is preferable to gradually diminish the motion field close to the mask boundaries. The resulting video features softer motion transitions, and is better suited for elements with no sharp boundaries such as fluids, smoke, and fire. We achieve this effect by placing artificial anchor points with zero displacement around the mask and including them in the interpolation of the displacement field (Section 3.1.3).

## 5 RESULTS AND EVALUATION

In Figures 8 and 9 we present the result of using our system on different types of input images, ranging from color-patterned animals to tiled floors and dotted dresses. Please note that it is hard to evaluate our results by looking at individual frames, and we urge the reader to watch these and many more videos in the supplemental website. Some images in Figure 9 are actually videos that can be played when this file is opened in *Acrobat Reader*.

### 5.1 User Studies

We compare our method quantitatively to those of Holynski et al. [2020] and Endo et al. [2019] via a user study. To this end, we collected a set of 102 images from a stock image website[2], the majority of which contain man-made objects, humans, and animals, and the rest consist of natural scenes with fluid elements such as water or smoke, to match the kind of inputs the other methods were trained on. From this dataset, we randomly sampled 25 images and applied our pipeline, as well as the network from the official implementation of [Endo et al. 2019]. The authors of [Holynski et al. 2020] kindly ran their model on this dataset and provided the output videos. We refer to this dataset as *Diverse Images*. We added a second set consisting of all 25 images presented on the project page of Holynski et al. [2020][3], which we will refer to as *Fluid Elements*, to get a test set of 50 images.

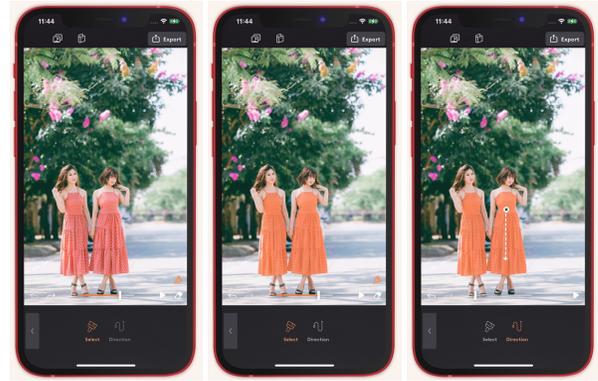

Fig. 7. *Our mobile app UI.* We present the user with a masking tool (center) (implementing a Paint Selection [Liu et al. 2009] algorithm for ease of masking) and a tool to select the direction of motion (right). Image credit: © Bùi Thanh Tâm

Table 1. *User study.* Users were asked for each video: (a) *"Is this video aesthetically pleasing?"* and (b) *"What is the amount of visually disturbing artifacts in the video?"* (Scale [1-5], higher is better). We present the mean scores, calculated separately on the *Fluid Elements* videos from Holynski et al. [2020] and on a set of *Diverse Images* chosen by us. We outperform the method of Endo et al. on both datasets. The method of Holynski et al. excels on their use case, but cannot handle images without objects that undergo fluid motion such as water or smoke. Adding the CRF formulation improves our results for *diverse images*, and does not make a difference for *fluid elements*. (The difference between pairs marked with * and ** is not statistically significant.) [†] For many of the images in the *Diverse* set, the method of Holynski et al. gracefully degrades to generating no motion whatsoever. We discard these results, as they are still images and not looping videos.

|  | **Diverse Images** | | **Fluid Elements** | |
| --- | --- | --- | --- | --- |
| **Method** | Aesthetics | Artifacts | Aesthetics | Artifacts |
| Endo et al. | 1.9 | 1.8 | 2.3 | 2.1 |
| Holynski et al. | 3.4[†] | 3.7[†] | **4.3** | **4.2** |
| Ours w/o CRF | 3.5 | 3.5 | 3.3* | 3.3** |
| Ours | **3.8** | **3.8** | 3.2* | 3.2** |

---

[2]www.unsplash.com
[3]https://eulerian.cs.washington.edu

142:8 • Halperin, T. et al

In addition to the outputs of the three methods we also generated an *ablated* output, where we skipped the CRF step and directly assigned each pixel with the nearest label from the 1D step. This worked quite well in some cases, in particular for highly regular patterns, while eliminating the most time consuming component of our system.

We presented the videos to a group of 71 users. We shuffled the videos and each user was shown all four versions (*Endo et al., Holynski et al., Ours without CRF, Ours with CRF*) of the videos created from 10 randomly selected images. They were asked to rate the degree at which the video is aesthetically pleasing on a 1-5 scale, and the amount of visually disturbing artifacts in the video on the same scale. We separately analyse the two types of inputs: *Diverse Images* and *Fluid Elements*. We note that the other methods were trained on images similar to the latter, so comparing to them on the former is somewhat unfair since this kind of data lies outside of the distribution of images they were trained on. Nevertheless, this emphasizes the uniqueness of our method, which can create surreal cinemagraphs containing motion which cannot appear in the real world.

We summarize the user study in Table 1, and show a qualitative comparison in Figure 8. The difference between conditions is statistically significant (Kruskal-Wallis test, $p < 10^{-91}$). We use *Tukey's honest significant difference procedure* between all pairs, properly taking into account multiple hypothesis testing. All differences between our result and other methods in Table 1 are statistically significant ($p < 0.04$ for all pairs, and usually much smaller), except for the artifacts question in the *Diverse Image* set when comparing our method to that of [Holynski et al. 2020]. We outperform the method of Endo et al. on both datasets. The method of Holynski et al. excels in their use case, but cannot handle images that do not contain objects with fluid motion such as water or smoke. All the videos from the user study are available on the supplementary website.

### 5.2 Synthetic Experiments

To verify the correctness of our method and to break down the relative contribution of each element of the algorithm we ran our system on a set of synthetic tests, evaluating the robustness with respect to increasing levels of pixel noise, as well as to different types of local and perspective deformations. We used the benchmark introduced in [Lukáč et al. 2017], which consists of synthetic images of symmetric patterns with increasing levels of spatial and pixel value noise. We test on the 'translation' subset which is the most relevant to us. (Other patterns such as reflections fit less to our scenario.) The results are presented on the supplementary website.

### 5.3 Mobile App

We created an optimized implementation in the form of a mobile app showcasing this algorithm[4]. Users can pick an image, draw a mask and indicate the desired direction of motion. We present screenshots of the app in Figure 7, and include workflow examples in our supplementary video. We conducted an *A/B test* inside the app, where 140,000 users manually created cinemagraphs by drawing sparse motion vectors to interpolate, and the other 140,000 used our tool instead. The test lasted for three months in which over 350,000 cinemagraphs were exported. Our method showed a 15% increase in usage and a 20% increase in the number of exports per session. We show a workflow comparison between the variants in our supplementary video.

---

[4]https://apps.apple.com/us/app/motionleap-by-lightricks/id1381206010

Table 2. *Comparison to previous methods.* **Time** is the total computation time (in seconds) to generate an output sequence of 80 frames. **Interact** - whether direction of motion is controllable. $^{\dagger}$Endo et al. [2019] allow indirect user control over motion direction via codebook sampling, contributing an average of 7 seconds to the total duration. Apart from that, the other methods are fully automatic, while ours requires users to manually indicate the moving areas. **Video** - whether real footage is needed for training. This requirement also limits the use cases to realistic motion.

| Method | Time | Hardware | Interact | Video |
|---|---|---|---|---|
| Endo et al. | $11^{\dagger}$ sec | GTX 1080 Ti GPU | Yes$^{\dagger}$ | Yes |
| Holynski et al. | 20 sec | Titan Xp GPU | No | Yes |
| Ours | 1 sec | iPhoneXS CPU | Yes | No |

### 5.4 Running Time

We run the entire algorithm on a downsampled version of the image, where the long side of the mask is 300 pixels. After obtaining a flow field we upsample it back to the original resolution of the image and generate the video in full resolution. The most time consuming part of the algorithm is the dense CRF optimization. Its running time is linear in the number of pixels and quadratic in the number of labels, thus it is essential to keep the number of labels as small as possible. For CRF optimization, we use a python wrapper[5] of the original C++ code [Krähenbühl and Koltun 2013]. We run 10 CRF iterations in all experiments. This number was chosen empirically, as we noticed almost no change in labeling with additional iterations. The entire pipeline takes 4-15 seconds for a typical image, with our unoptimized python version, depending on how many labels were used.

We experimented with replacing our VGG-based pixel difference with either $L_2$ pixel difference or pixel descriptors such as HOG or SIFT [Lowe 2004]. However, the former resulted in inferior accuracy while the latter was too slow on mobile devices (more than 20x slower).

The total running time of the optimized algorithm in our mobile app is 0.5-1 seconds on an iPhone XS. This speedup mostly stems from parallelizing the CRF step which consumes the majority of the running time. We utilize multiple CPU cores, and make massive use of SIMD (Single Instruction Multiple Data). To further reduce the running time of the mobile app, the unmasked pixels do not participate in the CRF optimization at all. Excluding the CRF step, the entire algorithm takes 200ms (with results similar to our ablated variant). We show a running time comparison in Table 2.

---

[5]https://github.com/lucasb-eyer/pydensecrf

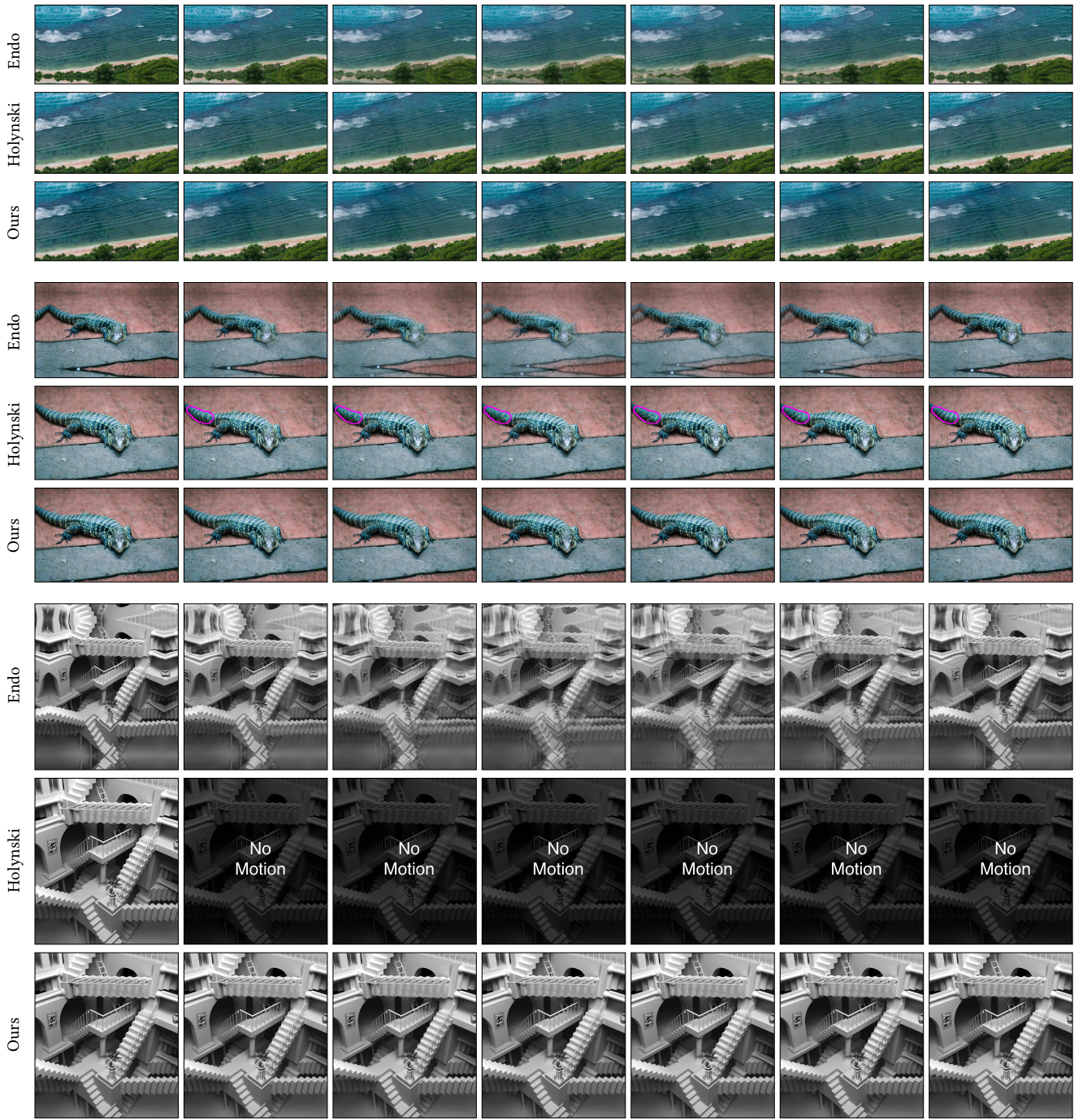

Fig. 8. *Qualitative comparison with previous methods.* **Top:** Fluid motion from the project page of Holynski et al. [2020]. **Middle:** The motion produced by Holynski et al. in this case covers only a tiny area of the image (delineated in the image in magenta), and neither the area nor the direction is controllable by the user. **Botton:** The method of Holynski et al. produces no motion here. In all the images, the method of Endo et al. produces motion with noticeable artifacts. Videos can be found on the supplementary website. Image credits: © Fotocrisis, John Hanusek

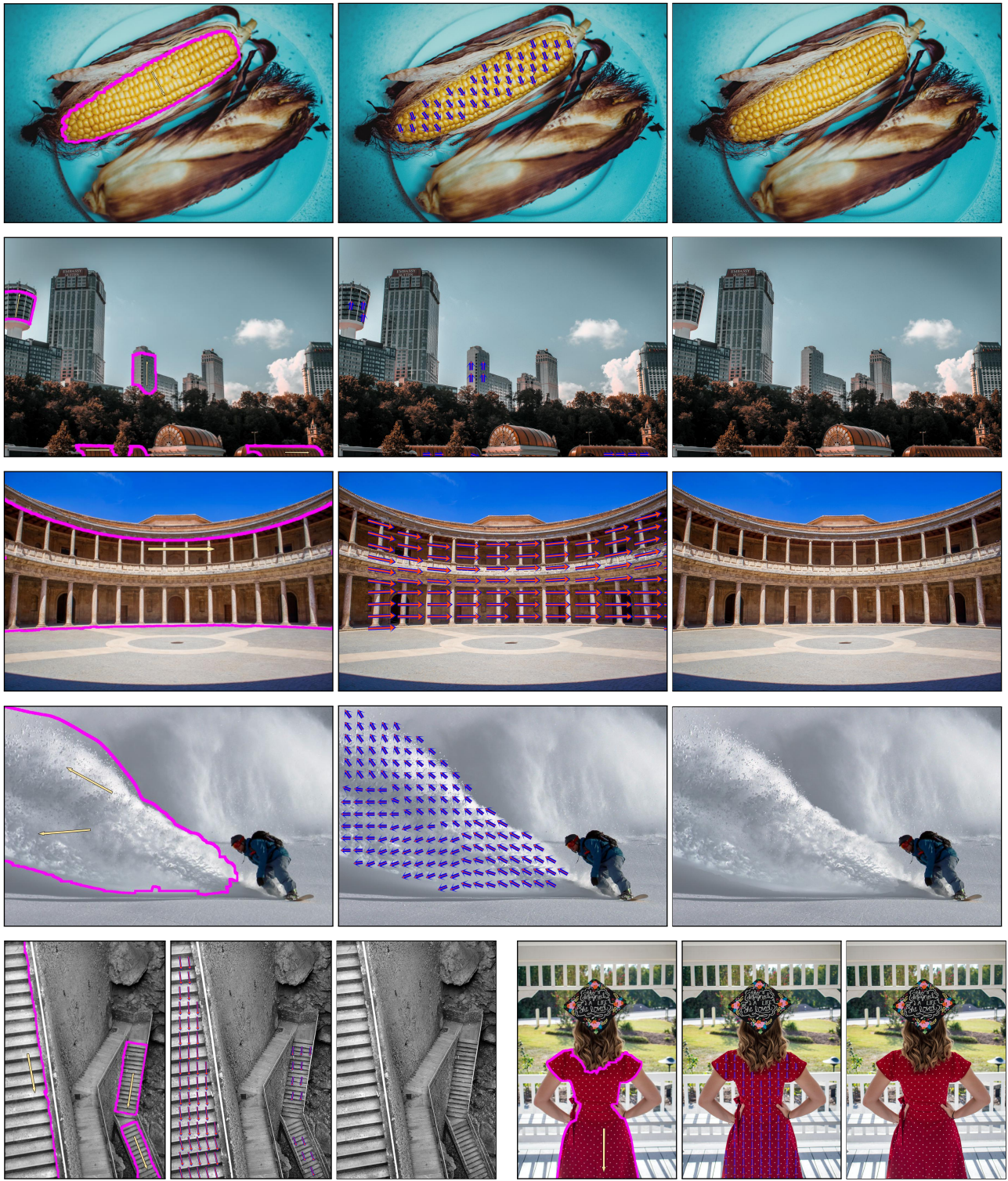

Fig. 9. *Typical output, man-made and natural scenes.* Our algorithm works on periodic patterns but also produces appealing results for natural images with less structure. For each example we show the input in the form of a mask and main direction(s), the output of the CRF as superimposed arrows, and the produced cinemagraph. **Please open in Acrobat Reader to see the embedded videos.** Image credits: © Aleks Marinkovic, Liz Weddon, Clark Van Der Beken, Johannes Waibel, Daniele Pietrobelli, Monica Smith

## 6 LIMITATIONS AND DISCUSSION

Our method assumes the patterns in the area of interest to be fairly regular. When a pattern breaks this assumption, the generated looping video may not follow the pattern. One failure point is that the 1D repetition detection step may not latch on the pattern when it changes abruptly; another is that the CRF labels may only partially cover the span of desired motions in the image. A third point of failure may be the CRF itself that assumes a smooth output motion field.

We can mitigate the second type of failure, where more CRF labels are needed to precisely describe the 'motion' of the pattern, by increasing the span of labels we consider in the CRF optimization, for instance by setting a larger maximal angular deviation (Section 3.1.2). However, this comes at a price in terms of the running time and noisiness of the CRF optimization. Although in some cases we found that increasing this number yielded more accurate results, we keep it constant as there was no simple heuristic to predetermine this number based on the content of an image.

Examples of failure cases can be found on the supplementary website. We ran our algorithm on the "translation" subset of the synthetic image set by [Lukáč et al. 2017]. Similarly to their results, our algorithm can handle additive pixel noise, local image distortions or reasonable perspective distortion, but fails to accurately align the motion with the image after strong perspective warping.

Another limitation of our algorithm is dealing with rotational motion. We cannot solve for rotational motion in a single CRF optimization as it does not fit our label set construction with its limited range of angles. Instead, the user can draw multiple strokes (section 4.2) around the object. This needs to be done carefully in order for the different areas of motion to fit together, and requires usually a few iterations to get right. We show such examples in the supplementary materials.

Our method is not sensitive to small changes in the input direction. When the algorithm latches on a pattern, rotating the input direction by a small angle (within the tolerance of the CRF labels) will not result in a significant difference in the output video. Rotating by a large angle will make the algorithm latch on a different pattern (Figure 10).

When animating fluids, we occasionally observe ghosting artifacts that appear due to alpha blending. Our initial attempts to eliminate these were not successful, and we left it for future work.

## 7 CONCLUSIONS

Our proposed algorithm for the generation of cinemagraphs from still images was shown to be:

(a) *Performant and reliable*, as proven by its interactive implementation inside a mobile app, successfully deployed on various devices and used by thousands of users.
(b) *Flexible*, as it can handle automatic or user-defined input of direction of motion, and can animate images with multiple regions moving in different directions. It can also produce both natural-looking videos (smoke, snow, water), as well as interesting surreal animations such as those presented in the accompanying video and the supplemental website.

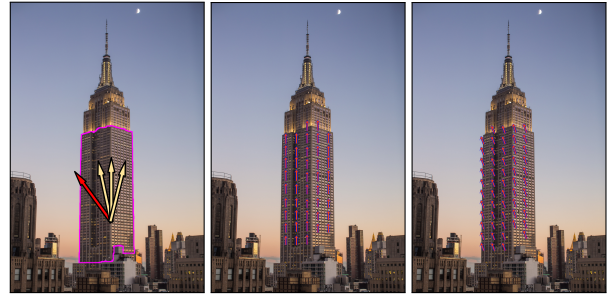

Fig. 10. *Robustness to input direction.* Our algorithm is robust to minor perturbations of the selected initial direction. We show varying input directions (left, yellow arrows), all resulting in almost identical output (middle). Strongly varying the input direction (left, red arrow) will create motion in a different direction of repetition of the pattern (right). Image credit: © Ben Dumond

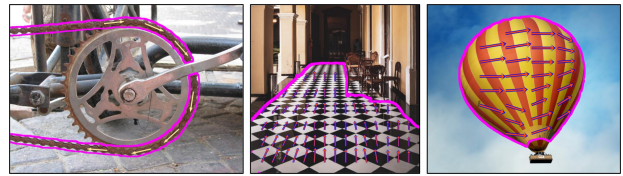

Fig. 11. *Limitations.* **Left:** To achieve a rotating cinemagraph the user needs to carefully draw multiple motion directions. **Center:** The algorithm may fail to detect repetitions under strong perspective distortion, as the patterns appear distorted and in multiple scales. **Right:** Our CRF step depends on the labels obtained at the first step; if the correct labels are not available it will not recover. In the balloon example the repetition length varies greatly perpendicular to the main direction, thus proper short labels are not available in the lower half of the image. *Videos can be found on the supplementary website.* Image credits: © Republica, Antonio Alcántara, Faye Cornish

(c) *Aesthetically pleasing*, as seen quantitatively in our user studies.

We currently use a model-based approach for motion estimation rather than a data-driven one, getting around the limitation of not having any training data for our task. Given the recent developments in unsupervised optical flow estimation (e.g. [Liu et al. 2020]), it would be interesting to also try a *data-driven approach to training for flow generation in an unsupervised manner*.

Another interesting direction would be to explicitly *incorporate 3D priors and 3D knowledge* about the scene. This can improve visual quality, and also produce better sequences when using challenging videos as input.

Lastly, in our current implementation the user is required to draw a mask. While we do not want to decide on a mask completely automatically (since we want the user to exert artistic control), it would be interesting to *automatically detect several possible masks* and suggest them to the user to facilitate exploration.

We invite the reader to download our app and start making beautiful cinemagraphs!

## ACKNOWLEDGMENTS

We would like to thank Gal Nachmana for the valuable discussions, Paz Bendek for the mobile app performance data, Matt Veysberg for producing the supplementary video, and Kaley Halperin for narrating it. We thank Aleksander Holynski for providing results of his method.